\pdfoutput=1

\documentclass{ws-procs11x85}
\usepackage{ws-procs-thm}   

\usepackage[draft]{fixme}
\fxsetup{inline,nomargin,theme=color}

\begin{document}

\title{CheXclusion: Fairness gaps in deep chest X-ray classifiers}

\author{Laleh Seyyed-Kalantari$^{1,2}$\footnote{Corresponding author email: \texttt{laleh@cs.toronto.edu}}, Guanxiong Liu$^{1,2}$, Matthew McDermott$^3$, Irene Y. Chen$^3$, Maryzeh Ghassemi$^{1,2}$}

\address{$^1$Computer Science, University of Toronto, Toronto, Ontario, Canada}

\address{$^2$Vector Institute, Toronto, Ontario, Canada}

\address{$^3$Electrical Engineering and Computer Science, Massachusetts Institute of Technology,\\
Cambridge, MA USA}

\begin{abstract}
Machine learning systems have received much attention recently for their ability to achieve expert-level performance on clinical tasks, particularly in medical imaging. 
Here, we examine the extent to which state-of-the-art deep learning classifiers trained to yield diagnostic labels from X-ray images are biased with respect to \textit{protected attributes}.
We train convolution neural networks to predict 14 diagnostic labels in 3 prominent public chest X-ray datasets: MIMIC-CXR, Chest-Xray8, CheXpert, as well as a multi-site aggregation of all those datasets. We evaluate the \textit{TPR disparity} -- the difference in true positive rates (TPR) -- among different protected attributes such as patient sex, age, race, and insurance type as a proxy for socioeconomic status. 
We demonstrate that TPR disparities exist in the state-of-the-art classifiers in all datasets, for all clinical tasks, and all subgroups. A multi-source dataset corresponds to the smallest disparities, suggesting one way to reduce bias.
We find that TPR disparities are not significantly correlated with a subgroup's proportional disease burden.  As clinical models move from papers to products, we encourage clinical decision makers to carefully audit for algorithmic disparities prior to deployment. Our code can be found at,  \texttt{https://github.com/LalehSeyyed/CheXclusion}.
\end{abstract}

\keywords{fairness, medical imaging, chest x-ray classifier, computer vision.}

\copyrightinfo{\copyright\ 2020 The Authors. Open Access chapter published by World Scientific Publishing Company and distributed under the terms of the Creative Commons Attribution Non-Commercial (CC BY-NC) 4.0 License.}
\section{Introduction}
Chest X-ray imaging is an important screening and diagnostic tool for several life-threatening diseases, but due to the shortage of radiologists, this screening tool cannot be used to treat all patients.~\cite{rimmer_radiologist_2017, shortage2015}
Deep-learning-based medical image classifiers are one potential solution, with significant prior work targeting chest X-rays specifically,\cite{wang_chestx-ray8:_2017,yao_learning_2017} leveraging large-scale publicly available datasets,\cite{wang_chestx-ray8:_2017, johnson_mimic-cxr:_2019, irvin_chexpert:_2019} and demonstrating radiologist-level accuracy in diagnostic classification.\cite{rajpurkar_chexnet:_2017, rajpurkar_deep_2018, irvin_chexpert:_2019}

Despite the seemingly clear case for implementing AI-enabled diagnostic tools,~\cite{vect} moving such methods from paper to practice require careful thought.~\cite{ghassemi2019practical}  Models may exhibit disparities in performance across protected subgroups, and this
could lead to different subgroups receiving different treatment.~\cite{chen2019can} During evaluation, machine learning algorithms usually optimize for, and report performance on, the general population rather than balancing accuracy on different subgroups. While some variance in performance is unavoidable, mitigating any systematic bias against protected subgroups may be desired or required in a deployable model. 

In this paper, we examine whether state-of-the-art (SOTA) deep neural classifiers trained on large public medical imaging datasets are fair across different subgroups of \textit{protected attributes}. We train classifiers on 3 large, public chest X-ray datasets: MIMIC-CXR,~\cite{johnson_mimic-cxr:_2019} CheXpert,\cite{irvin_chexpert:_2019} Chest-Xray8 \cite{wang_chestx-ray8:_2017}, as well as an additional datasets formed of the aggregation of those three datasets on their shared labels. In each case, we implement chest X-ray pathology classifiers via a deep convolutional neural network (CNN) chest X-ray images as inputs, and optimize the multi-label probability of 14 diagnostic labels simultaneously.
Because researchers have observed health disparities with respect to race,~\cite{kawachi2005health} sex,~\cite{hoffmann2001girl} age,~\cite{walter2019age} and socioeconomic status,~\cite{kawachi2005health} we extract structural data on race, sex, and age; we also use insurance type as an imperfect proxy~\cite{chen2019can} for socioeconomic status. To our knowledge, we are the first to examine whether SOTA chest X-ray pathology classifiers display systematic bias across race, age, and insurance type. 

We analyze equality of opportunity~\cite{hardt_equality_2016} as our fairness metric based on the needs of the clinical diagnostic setting.
In particular, we examine the differences in true positive rate (TPR) across different subgroups per attributes. A high TPR disparity indicates that sick members of a protected subgroup would \emph{not} be given correct diagnoses---e.g., true positives---at the same rate as the general population, even in an algorithm with high overall accuracy. 

We find three major findings: First, that there are indeed extensive patterns of bias in SOTA classifiers, shown in TPR disparities across datasets. Secondly, the disparity rate for most attributes/ datasets pairs is not significantly correlated with the subgroups' proportional disease membership. These findings suggest that underrepresented subgroups could be vulnerable to mistreatment in a systematic deployment, and that such vulnerability may not be addressable simply through increasing subgroup patient count. Lastly, we find that using the multi-source dataset which combines all the other datasets yields the lowest TPR disparities, suggesting using multi-source datasets may combat bias in the data collection process.
As researchers increasingly apply artificial intelligence and machine learning to precision medicine, we hope that our work demonstrates how predictive models trained on large, well-balanced datasets can still yield disparate impact. 

\section{Background and Related Work}
\label{sec:Related work}

\textbf{Fairness and Debiasing.} Fairness in machine learning models is a topic of increasing attention, spanning sex bias in occupation classifiers,\cite{de2019bias} racial bias in criminal defendant risk assessments algorithms,\cite{chouldechova_fair_2016} and intersectional sex-racial bias in automated facial analysis.\cite{shade18} Sources of bias arise in many different places along the classical machine learning pipeline. For example, input data may be biased, leaving supervised models vulnerable to labeling and cohort bias.\cite{shade18} Minority groups may also be under-sampled, or the features collected may not be indicative of their trends.\cite{chen_why_2018} There are several conflicting definitions of fairness, many of which are not simultaneously achievable.\cite{kleinberg_inherent_2016} The appropriate choice of a disparity metric is generally task dependent, but balancing error rates between different subgroups is a common consideration, \cite{chouldechova_fair_2016,hardt_equality_2016} with equal accuracy across subgroups being a popular choice in medical settings.\cite{srivastava2019mathematical} In this work, we consider the equality of opportunity notion of fairness and evaluate the rate of correct diagnosis in patients across several protected attribute groups.

\textbf{Ethical Algorithms in Health.} Using machine learning algorithms to make decisions raises serious ethical concerns about risk of patient harm.\cite{NEJMp1714229} Notably, biases have already been demonstrated in several settings, including racial bias in the commercial risk score algorithms used in hospitals,\cite{racehealth19} or an increased risk of electronic health record (EHR) miss-classification in patients with low socioeconomic status.\cite{biasEHR} It is crucial that we actively consider fairness metrics when building models in systems that include human and structural biases.


\textbf{Chest X-Ray Classification.}
With the releases of large public datasets like Chest-Xray8,~\cite{wang_chestx-ray8:_2017} CheXpert,~\cite{irvin_chexpert:_2019} and MIMIC-CXR,~\cite{johnson_mimic-cxr:_2019} many researchers have begun to train large deep neural network models for chest X-ray diagnosis. \cite{rajpurkar_deep_2018,yao_learning_2017,irvin_chexpert:_2019, akbarian2020evaluating} Prior work\cite{rajpurkar_deep_2018} demonstrates a diagnostic classifier trained on Chest-Xray8 can achieve radiologist-level performance. Other work on CheXpert~\cite{irvin_chexpert:_2019} reports high performance for five of their diagnostic labels. To our knowledge, however, no works have yet been published which  
examined whether any of these algorithms display systematic bias over age, race and insurance type (as a proxy of socialeconomic status).

  
\section{Data} 
\label{sec:Dataset}
We use three public chest X-ray radiography datasets described in Table \ref{tbl:datasets}: MIMIC-CXR (CXR),\cite{johnson_mimic-cxr:_2019} CheXpert (CXP),\cite{irvin_chexpert:_2019}  Chest-Xray8 (NIH).\cite{wang_chestx-ray8:_2017} Images in CXR, CXP, and NIH are associated with 14 diagnostic labels (see Table~\ref{Table:Acc}). We combine all non-positive labels within CXR and CXP (including ``negative'', ``not mentioned'', or ``uncertain'') into an aggregate ``negative'' label for simplicity, equivalent to ``U-zero'' study of `NaN' label in CXP. In CXR and CXP, one of the 14 labels is ``No Finding'', meaning no disease has been diagnosed for the image and all the other 13 labels are 0. Of the 14 total disease labels, only 8 are shared amongst all 3 datasets. Using these 8 labels, we define a multi-site dataset (ALL) that consists of the aggregation of all images in CXR, CXP, and NIH defined over this restricted label schema.

These datasets contain protected subgroup attributes, the full list of which includes sex (Male and Female), age (0-20, 20-40, 40-60, 60-80, and 80-), race (White, Black, Other, Asian, Hispanic, and Native) and insurance type (Medicare, Medicaid, and Other). These values are taken from the structured patient attributes. NIH, CXP, and ALL only have the patient sex and age, while CXR also has race and insurance type data (excluding around 100,000 images). 


\begin{table}[h]
\tbl{Description of chest X-ray datasets, MIMIC-CXR (CXR), \cite{johnson_mimic-cxr:_2019} CheXpert (CXP),\cite{irvin_chexpert:_2019} Chest-Xray8 (NIH).\cite{wang_chestx-ray8:_2017} and their aggregation on 8 shared labels (ALL). 
Here, the number of images, patients, view types, and the proportion of patients per subgroups of sex, age, race, and insurance type are presented. `Frontal' and `Latral' abbreviate frontal and lateral view, respectively. Native, Hispanic, and Black denote self-reported American Indian/Alaska Native, Hispanic/Latino, and Black/African American race respectively.}
{
\begin{tabular}{@{}llrrrr@{}}
\toprule
Subgroup & Attribute & \textbf{CXR}~\cite{johnson_mimic-cxr:_2019}& \textbf{CXP}~\cite{irvin_chexpert:_2019} & \textbf{NIH}~\cite{wang_chestx-ray8:_2017} & \textbf{ALL} \\ \colrule
& \# Images &  371,858 & 223,648 & 112,120 & 707,626\\
& \# Patients  & 65,079 & 64,740 & 30,805 & 129,819 \\ 
& View &  Frontal/Lateral & Frontal/Lateral & Frontal & Frontal/Lateral \\ \colrule
sex & Female & 47.83\% & 40.64\% & 43.51\% & 44.87\%\\
       & Male & 52.17\% & 59.36\% & 56.49\% & 55.13\% \\ \colrule
Age & 0-20 & 2.20\% & 0.87\% & 6.09\% & 2.40\%\\
& 20-40 & 19.51\% & 13.18\% & 25.96\% & 18.53\%\\
& 40-60 & 37.20\% & 31.00\% & 43.83\% & 36.29\%\\
& 60-80 & 34.12\% & 38.94\% & 23.11\% & 33.90\%\\
& 80+ & 6.96\% & 16.01\% & 1.01\% & 8.88\%\\ \colrule
Race & Asian & 3.24\% & --- & --- & --- \\
& Black & 18.59\% & --- & ---& ---\\
& Hispanic & 6.41\% & --- & --- & ---\\
& Native & 0.29\% & --- & --- & ---\\
& White & 67.64\% & --- & --- & ---\\
& Other & 3.83\% & --- & --- & ---\\ \colrule
Insurance & Medicare & 46.07\% & --- & --- & ---\\
& Medicaid & 8.98\% & --- & --- & ---\\
& Other & 44.95\% & --- & --- & ---\\ \botrule
\end{tabular}}\label{tbl:datasets}
\end{table}

\section{Methods} 
\label{sec:Methods}

We implement CNN-based models to classify chest X-ray images into 14 diagnostic labels. We train separate models for CXR,\cite{johnson_mimic-cxr:_2019} CXP,\cite{irvin_chexpert:_2019} NIH  \cite{wang_chestx-ray8:_2017} and ALL and explore their fairness with respect to patient sex and age for all 4 datasets as well as race and insurance type for CXR. 

\subsection{Models}

We initialize a 121-layer DenseNet\cite{huang_densely_2017} with pre-trained weights from ImageNet\cite{imagenet_cvpr09} and train multi-label models with a multi-label binary cross entropy loss. The 121-layer DenseNet was used as it produced the best results in prior studies \cite{irvin_chexpert:_2019, rajpurkar_deep_2018} . We use a 80-10-10 train-validation-test split with no patient shared across splits. We resize all images to $256 \times 256$ and normalize via the mean and standard deviation of the ImageNet dataset.\cite{imagenet_cvpr09} We apply center crop, random horizontal flip and random rotation, as some of the images maybe flipped or rotated within the dataset. The initial degree of random rotation is chosen by hyperparameter tuning.
We use Adam \cite{adam} optimization with default parameters, and decrease the learning rate (LR) by a factor of 2 if the validation loss does not improve over three epochs; we stop learning if validation loss does not improve over 10 epochs. Thus the ultimate  number of epochs for training each model is varied based on the early stop condition.  
For NIH, CXP and CXR we first tune models to get the highest average area under the receiver operating characteristic curve (AUC) over 14 labels by fine tuning the LR. For the best achieved model, we fine tune the degree of random rotation data augmentation from the set of 7, 10 and 15 and select the best model. Following this, best initial LR is 0.0005 for CXR and NIH where it is achieved as 0.0001 for CXP. Also, best initial degree for random rotation data augmentation is 10 for NIH and 15 for the CXR and CXP.
For training on ALL, we use the majority vote of the best hyperparameters per individual dataset (e.g. 0.0005 initial LR and 15 degree random rotation). We then, fix the hyperparameters of the best model and train four extra models with the same hyperparameters but different random seeds
between 0 to 100, per dataset. We report all the metrics based on the mean and 95\% confidence intervals (CI) achieved over five studies per dataset. We choose batch size of 48 to use the maximum memory capacity of the GPU, for all datasets except NIH where we choose 32 similar to prior work \cite{rajpurkar_deep_2018}. 
The output of the network is an array of 14 numbers between 0 and 1 indicating the probability of each disease label.
The binary prediction threshold per disease is chosen to maximize the F1 score measure on the validation dataset. We train models using a single NVIDIA GPU with 16G of memory in approximately 9, 20, 40, and 90 hours for NIH, CXP, CXR, and ALL, respectively.

\subsection{Classifier Disparity Evaluation}
\label{sec:Formulation}
Our primary measurement of bias is \emph{TPR disparity}. For example, given a binary subgroup attribute such as sex (which in our data we classify as either `male' or `female'), we mimic prior work\cite{de2019bias} and define the TPR disparity per diagnostic label $i$ as simply the TPR of label $i$ restricted to female patients minus that for male patients. More formally, letting $g$ be the binary subgroup attribute, we define $\operatorname{TPR}_{g, i}=P[\hat{Y_{i}}=y_{i} | G=g, Y_{i}=y_{i}]$, and the  TPR disparity as, $\operatorname{Gap}_{g,i}=\operatorname{TPR}_{g, i}-\operatorname{TPR}_{\neg g, i}$. For non-binary attributes $S_1, \ldots, S_N$, we use the difference between a subgroup's TPR and the median of all TPRs to define TPR disparity of the $j$th subgroup for the $i$th label as, $\operatorname{Gap}_{S_j, i}=\operatorname{TPR}_{S_j, i}- \mathrm{Median}(\mathrm{TPR}_{S_{1}, i},..,\mathrm{TPR}_{S_k, i})$.

\section{Experiments}
\label{Sec:Experiments}
First, we demonstrate that the classifiers we train on all datasets reach near-SOTA level performance. This motivates using them to study fairness implications, as we can be confident any problematic disparities are not simply reflective of poor overall performance. 
Next, we explicitly test these classifiers for their implications on fairness. We target two investigations:
\begin{enumerate}
\item \textbf{TPR disparity:} We quantify the TPR disparity per subgroup/disease for sex and age across all 4 datasets, and due to data availability for race and insurance type on CXR.  

\item \textbf{TPR disparity in proportion to membership:} We investigate the distribution of the positive patient proportion per subgroup $S_{j}$ and label $y_{i}$ (which is given by $P[S=S_{j} | Y= y_{i}]$) and the effect on TPR disparities.
Prior work on chest X-ray diagnosis prediction has suggested data imbalance can explain sex TPR disparities\cite{larrazabal2020gender} while work in other domains illustrates that disparities in small or vulnerable subgroups could be propagated if put into practice,\cite{de2019bias, hashimoto_fairness_2018} and these experiments are meant to probe that hypothesis.
\end{enumerate}

\section{Results}

\begin{table}[h]
\tbl{The AUC for chest X-ray classifiers trained on CXP, CXR, NIH, and ALL averaged over 5 runs $\pm$95\%CI, where all runs have same hyperparameters but different random seed. 
(`Airspace Opacity'\cite{johnson_mimic-cxr:_2019} and `Lung Opacity' \cite{irvin_chexpert:_2019} denote the same label.)}
{\begin{tabular}{@{}llccccc@{}}
\toprule
& \textbf{Label (Abbr.)}  & \textbf{CXR}               & \textbf{CXP}           & \textbf{NIH} & \textbf{ALL} \\ \colrule
& Airspace Opacity (AO)   & 0.782 $\pm$  0.002         & 0.747  $\pm$  0.001    & --- & --- \\
& Atelectasis (A)         & 0.837 $\pm$  0.001         & 0.717  $\pm$  0.002    &  0.814  $\pm$  0.004 & 0.808  $\pm$  0.001\\ 
& Cardiomegaly (Cd)       & 0.828 $\pm$  0.002         & 0.855  $\pm$  0.003    & 0.915  $\pm$  0.002  & 0.856  $\pm$  0.001\\
& Consolidation (Co)      & 0.844 $\pm$  0.001         & 0.734  $\pm$  0.004    & 0.801  $\pm$  0.005 & 0.805  $\pm$  0.001\\
& Edema (Ed)              & 0.904 $\pm$  0.002         & 0.849  $\pm$  0.001    &  0.915  $\pm$  0.003 & 0.898  $\pm$  0.001 \\
& Effusion (Ef)           & 0.933 $\pm$  0.001         & 0.885  $\pm$  0.001    & 0.875  $\pm$  0.002 & 0.922  $\pm$  0.001\\
& Emphysema (Em)          & ---                        & ---                    & 0.897  $\pm$  0.002 & ---\\
& Enlarged Card (EC)      & 0.757 $\pm$  0.003         & 0.668  $\pm$  0.005    & ---                     & --- \\
& Fibrosis                & ---                        & ---                    & 0.788  $\pm$  0.007 & --- \\ 
& Fracture (Fr)           & 0.718 $\pm$  0.007         & 0.790  $\pm$  0.006    & --- & ---\\
& Hernia (H)              & ---                        & ---                    & 0.978  $\pm$  0.004& ---\\
& Infiltration (In)       & ---                        & ---                    & 0.717  $\pm$  0.004 & --- \\
& Lung Lesion (LL)        & 0.772 $\pm$  0.006         & 0.780  $\pm$  0.005    & ---& ---\\
& Mas (M)                 & ---                        & ---                    & 0.829  $\pm$  0.006 & ---\\
& Nodule (N)              & ---                        & ---                    & 0.779  $\pm$  0.006 & ---\\
& No Finding (NF)         & 0.868 $\pm$  0.001         & 0.885  $\pm$  0.001    & --- &  0.890  $\pm$  0.000 \\
& Pleural Thickening (PT) & ---                        & ---                    & 0.813 $\pm$  0.006 & ---\\
& Pleural Other (PO)      & 0.848 $\pm$  0.003         & 0.795  $\pm$  0.004    & --- & --- \\
& Pneumonia (Pa)          & 0.748 $\pm$  0.005         & 0.777  $\pm$  0.003    & 0.759  $\pm$  0.012  & 0.784  $\pm$  0.001\\
& Pneumothorax (Px)       & 0.903 $\pm$  0.002         & 0.893  $\pm$  0.002    & 0.879  $\pm$  0.005 & 0.904 $\pm$ 0.002 \\
& Support Devices (SD)    & 0.927 $\pm$  0.001         & 0.898  $\pm$  0.001    & --- & --- \\ \colrule
& \textbf{Average}        & \textbf{0.834 $\pm$ 0.001} &\textbf{0.805$\pm$0.001}&  \textbf{0.840 $\pm$ 0.001} & \textbf{0.859 $\pm$ 0.001}\\ 
 \botrule
\end{tabular}}\label{Table:Acc}
\end{table}

One potential reason that a model may be biased is because of poor performance, but we demonstrate that our models achieve near-SOTA classification performance. Table~\ref{Table:Acc} shows overall performance numbers across all tasks and datasets. Though results have non-trivial variability, we show similar performance to the published SOTA of NIH,\cite{rajpurkar_deep_2018} the only dataset for which a published SOTA comparison exists for all labels. Note that the published results for CXP\cite{irvin_chexpert:_2019} are on a private, unreleased dataset of only 200 images and 5 labels. Our results for CXP are on a randomly sub-sampled test set of size 22,274 images, so the numbers for this dataset are not comparable to the published results there.

\begin{figure*}[!htb]
    \centering
       \includegraphics[width=0.99\textwidth]{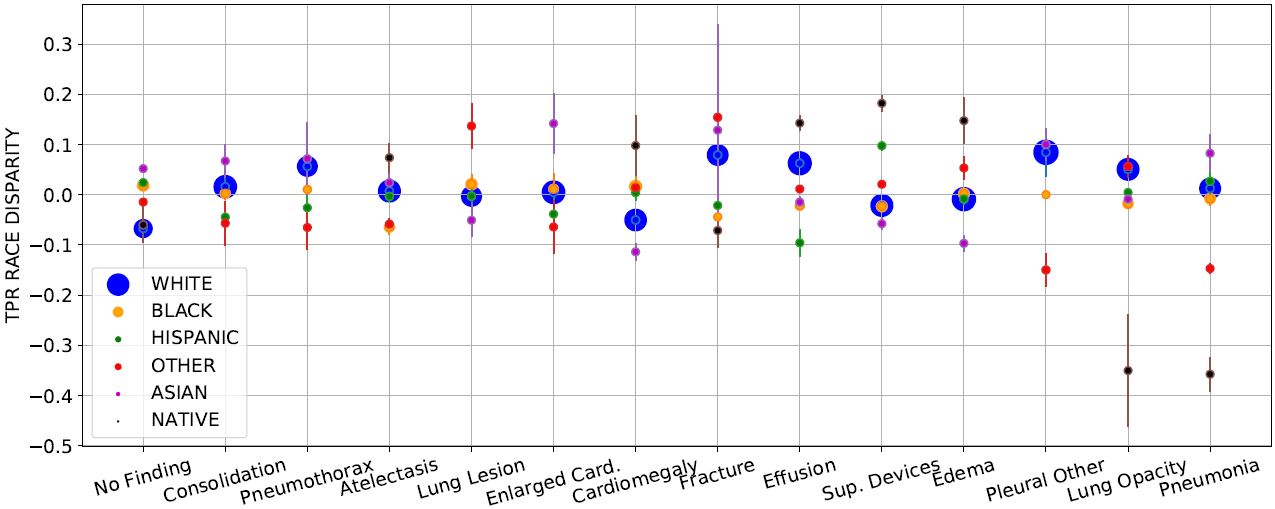}  
         \caption{
         The sorted distribution of TPR race disparity of CXR ($y$-axis) with label ($x$-axis). The scatter plot's circle area is proportional to group size. TPR disparities are averaged over five runs $\pm$95\%CI (
         shown with arrows). Hispanic patients are most unfavorable (highest count of negative TPR disparities, 9/13) whereas White patients are most favorable subgroup (9/13 zero or positive disparities). Labels `No Finding' (`NF') and `Pneumonia' (`Pa') have smallest (0.119) and largest (0.440) gap  between  least/most favorable subgroups. The average cross 14 labels gap is 0.226.}
       \label{MIMICRaceAll}
 \end{figure*}

\begin{table}[h]
\tbl{Disparities overview over attributes and datasets. 
We average per label gaps between the least and most favorable subgroup's TPR disparities per attributes/datasets to obtain the average cross-label gap. The
labels (full names on Table~\ref{Table:Acc}) that obtained the smallest and largest gaps are shown next to the average cross-label gap column, along with their gaps. We
summarize and label in columns the most frequent ``Unfavorable'' and ``Favorable'' subgroups count, which are the ones that experience TPRs disparities below or above the zero gap line.  See Section~\ref{sec:tpr_disparities} for more details.
}
{\begin{tabular}{@{}lllllll@{}}
\toprule

\textbf{Attribute} & \textbf{Dataset} & \multicolumn{3}{c}{\textbf{Average Cross-Label Gap}} & \textbf{Unfavorable} & \textbf{Favorable}\\
& & \textbf{Gap} & \textbf{Lowest} & \textbf{Greatest} & & \\ \colrule
Sex & ALL & \textbf{0.045} & Ef:0.001 & Pa:0.105        & Female (4/7)       & Male (4/7) \\    
& CXP &  0.062 & Ed:0.000 & Co:0.139       & Female (7/13)       & Male (7/13)  \\
& CXR  & 0.072 & Ed:0.011 & EC:0.151        & Female (10/13)      & Male (10/13)     \\
& NIH & 0.190 & M:0.001 &  Cd:0.393          & Female (8/14)       & Male (8/14) \\ \colrule
Age & ALL  & \textbf{0.215} & Ef:0.115 & NF:0.444       & 0-20 (5/7)        & 40-60,60-80(5/7)  \\
& CXR   & 0.245 & SD:0.091 & Cd:0.440        & 0-20, 20-40 (7/13)         & 60-80 (10/13)  \\ 
& CXP & 0.270 &SD:0.084& NF:0.604       & 0-20, 20-40, 80- (7/13)    & 40-60 (8/13)  \\
& NIH & 0.413 & In:0.188 & Em:1.00      & 60-80 (7/14)        & 20-40 (9/14) \\ \colrule
  Race & CXR  & 0.226  & NF:0.119 &  Pa:0.440         &  Hispanic (9/13)        & White (9/13) \\ \colrule
  Insurance & CXR & 0.100 & SD:0.021 &PO:0.190       & Medicaid (10/13)    & Other (10/13) \\
 \botrule
\end{tabular}}\label{Table:Summary}
\end{table}

\label{sec:Findings}   

\subsection{TPR Disparities}
\label{sec:tpr_disparities}
We calculate and identify TPR disparities and 95\% CI across all labels, datasets and attributes. We see many instances of positive and negative disparities, which can denote bias for or against of a subgroup, here referred to \emph{favorable} and \emph{unfavorable} subgroups. As an illustrative example Fig. \ref{MIMICRaceAll} shows the race TPR disparities distribution sorted by the the gap between least and most favorable subgroups per label. In a fair setting, all subgroups would have no appreciable TPR disparities, yielding a gap between least and most favorable subgroups within a label at `0'. Table~\ref{Table:Summary} shows the summary of the disparities in all attributes and datasets.  We note that the most frequent unfavorable subgroups are those with social disparities in the healthcare system, e.g., women and minorities, but no disease is consistently at the highest or lowest disparity. 
We show the average cross-label gap between and the labels of the least and most favorable subgroups per dataset and attributes. We count the number of time each subgroups experience negative disparities (unfavorable) and zero or positive disparities (favorable) across disease labels
and report the most frequent unfavorable and favorable subgroups by count in Table~\ref{Table:Summary}. For CXP and CXR, we exclude ``No Finding'' label in the count (counts are out of 13) as we want to check negative bias in disease labels.
Notably, the model trained on ALL has the smallest average cross- label gap between least/most favorable groups for sex and age. 
 
\subsection{TPR Disparity Correlation with Membership Proportion}
\label{sec:proportion}
We measure the Pearson correlation coefficients ($r$) between the TPR disparities and patients proportion per label across all subgroups/datasets. As we test multiple (33) hypotheses, (33 total comparisons amongst all protected attributes considered) with a desired significance level ($p<0.05$), then based on Bonferroni correction \cite{Bonferroni}, the statistical significance level for each hypothesis is $p<0.0015$ (0.05/33). The majority of correlation coefficients listed are positive, but the only statistically significant correlations are: race Other ($r$: 0.782, $p$: 0.0009) \& age subgroups, 20-40 ($r$: 0.766, $p$: 0.0013), 60-80 ($r$: 0.787, $p$: 0.0008) and 80- ($r$:0.858 , $p$: 0.0000) in CXR, age group 60-80 ($r$: 0.853, $p$: 0.0001) in CXP, and age group 60-80 ($r$: 0.936, $p$: 0.0006) in ALL.

\section{Summary and Discussion}
\label{sec:Discussion}

We present a range of findings on the potential biases of deployed SOTA X-ray image classifiers over the sex, age, race and insurance type attributes on models trained on NIH, CXP and CXR. We focus on TPR disparities similar to prior work,\cite{de2019bias} checking if the sick members of the different subgroups are given correct diagnosis at similar rates. 

Our results demonstrate several main takeaways. First, all datasets and tasks display nontrivial TPR disparities. These disparities could pose serious barriers to effective deployment of these models and invite additional changes in either dataset design and/or modeling techniques to ensure more equitable models. Second, using a multi-source dataset leads to smaller TPR disparities, potentially due to removing bias in the data collection process. Third, while there is occasionally a proportionality between protected subgroup membership per label and TPR disparity, this relationship is not uniformly true across datasets and subgroups. 

\subsection{Extensive Patterns of Bias}
We find that all datasets and tasks contain meaningful patterns of bias although no diseases are consistently at the highest or lowest disparity rates across all attributes and datasets. These disparities are present with respect to age and sex in all settings, with consistent subgroups (female, 0-20) showing consistently unfavorable outcomes. Note that in the case of the sex disparities, ``female'' patients are universally the least favored subgroup \emph{despite} the fact that the proportion of female patients is only slightly less than male patients in all 4 datasets.


We also observe TPR disparities with respect to the patient race and insurance type in the CXR dataset.  White patients, the majority, are the most favorable subgroup, where Hispanic patients are the most unfavorable. Additionally, bias exists against patients with Medicaid insurance, who are the minority population and are often from lower socioeconomic status. They are the most unfavorable subgroup with the model often providing incorrect diagnoses.

\subsection{Bias Reduction Through Multi-source Data}
Of the four datasets, the multi-source dataset led to the smallest disparities with respect to age and sex. Based on notions of generalizability in healthcare,~\cite{zech_confounding_2018,ghassemi2019practical} we hypothesize that this improvement stems from the combination of large datasets reducing data collection bias.

\subsection{Correlation Between TPR Disparities and Membership Proportion}
\label{sec:imbalance_data}
Although prior work has raised data imbalance as a potential cause of sex bias,~\cite{larrazabal2020gender} we observe TPR disparities are not often significantly correlated with disease membership. While we observe positive correlation between subgroups membership and TPR disparities, only 6 of 33 subgroups showed statistically significant correlation. By inspection, we identify diseases with the same patient proportion of a subgroup and completely different TPR disparities (e.g. `Consolidation', `Nodule' and `Pneumothorax' in NIH have 45\% Female, but the TPR disparities are in diverse range, -0.155, -0.079 and 0.047, respectively). Thus, having the same portion of images within all labels may not guarantee lack of bias.

\subsection{Discussion}
We identify subgroups that may experience more bias through the exploration of variance in TPR and FPR. Based on the equality of opportunity notion of fairness, a fair network should exhibit the same TPR on average among all subgroups regardless of how likely a subgroup may have a disease. Such an improvement would allow two patients with the same condition, but in different subgroups, to be diagnosed correctly and receive the same level of care. While we focused on some of the more obvious protected attributes, it is important to note that there are several other factors, subgroups, and attributes that we have not considered.

Identifying and eliminating disparities is particularly important as large datasets begin to be used by high-capacity neural models, but are based on highly skewed population, e.g., kidney injury prediction in a population that is 93.6\% male.\cite{tomavsev2019clinically} While chest X-ray images datasets are not sex-skewed, we note that the age, race and insurance type attributes are highly unbalanced, e.g., 67.6\% of patients are White, and only 8.98\% are under Medicaid insurance. Subgroups with chronic underdiagnosis are those who experience more negative social determinants of health, specifically, women, minorities, and those of low socioeconomic status. Such patients may use healthcare services less than others. In some groups, such a dataset skew can increase the risk of miss-classification.\cite{biasEHR}

Although ``de-biasing'' techniques \cite{amini_uncovering_2019, zhang_mitigating_2018} may reduce disparities, we should not ignore the important biases inherent in existent large public datasets. There are a number of reasons why dataset may induce disparities in algorithms, from imbalanced datasets to differences in statistical noise in each group (e.g. unmeasured predictive features) to differences in access to healthcare for patients of different groups.~\cite{chen_why_2018, kawachi2005health} For instance, an algorithm that can classify skin cancer \cite{esteva_dermatologist-level_2017} with high accuracy will not be able to generalize on different skin color if similar samples have not been represented enough in the trained dataset.\cite{shade18} Intentionally adjusting the datasets to reduce disparities in to protect minorities and the subgroups with high disparities is one potential option in dataset creation, though our analyses suggest that dataset membership cannot always ameliorate bias.

With the great promise of advanced models for clinical care, we caution that advanced SOTA models must be carefully checked for such biases as those we have identified. Disparities in small or vulnerable subgroups could be propagated\cite{hashimoto_fairness_2018} within the development of machine learning models. This raises serious ethical concerns\cite{NEJMp1714229} about the equal accessibility to the required medical treatment. Usually the SOTA classifiers are trained to provides high AUC or accuracy on the general population. However we suggest additionally applying rigorous fairness analyses before deployment. Clear disclaimers about the dataset collection process and potential resulting algorithmic bias could improve model assessment for clinical use.

\section{Limitations and Future Work}
\label{sec:Limitations and future work}

As SOTA deep learning diagnosis algorithms become more promising for medical screening, model bias investigation is essential. This work is a first step in quantifying these biases so that approaches for amelioration can be developed. However, important future work remains.

First, we note that across these models, our source of diagnostic labels for these images must be considered at best ``silver'' labels, as all currently existing public chest X-ray datasets use automatically deteremined labels based on natural language processing (NLP) techniques to extract labels from the radiology reports. These silver labels may be incorrect, in ways that could compound with observe biases or model errors, a risk that warrants further investigation. 
%
Additionally, we must consider the quality of the imaging devices themselves, the region of data collection, and the patient demographics at each hospital collection site. For instance, NIH was gathered from a hospital that covers more complicated cases, CXP contains more tertiary cases, and CXR was gathered from an emergency department, and prior literature has already shown that models are fully capable of taking advantage of such confounders.\cite{zech_confounding_2018} These challenges may affect both the label quality,\cite{lukeoakdenrayner_half_2019} and any patterns of bias in the labels, thereby affecting the resulting fairness metrics. Finally, exploration of existing de-biasing techniques, however limited, should also be undertaken over this modality to see if any of the problems we identified here can be resolved.



\section{Conclusion}
\label{sec:Conclusion}
While the development and deployment of machine learning models in a clinical setting poses exciting opportunities, great care must be taken to understand how existing biases may be exacerbated and propagated. We show the TPR disparity of SOTA chest X-ray pathology classifiers trained on 4 datasets, (MIMIC-CXR, ChestX-ray8, CheXpert, and aggregation of those three on shared labels) across 14 diagnostic labels. We quantify the TPR disparity across datasets along sex, age, race and insurance type. Our results indicate that high-capacity models trained on large datasets do not provide equality of opportunity naturally, leading instead to potential disparities in care if deployed without modification.



\section*{Acknowledgment}
We acknowledge the support of the Natural Sciences and Engineering Research Council of Canada (NSERC, funding number PDF-516984), Microsoft Research, CIFAR, NSERC Discovery Grant, and high performance computing platforms of Vector Institute. We also thank Dr. Alistair Johnson, Dr. Errol Colak  and  Grey Kuling for productive discussions.

\bibliographystyle{ws-procs11x85}
\bibliography{references}

\newpage
\appendix{Distribution of TPR Disparity per Attributes, Subgroups and Labels}

\label{Appendix bias}

\begin{figure*}[!htb]   
    \centering
      \includegraphics[width=0.99\textwidth]{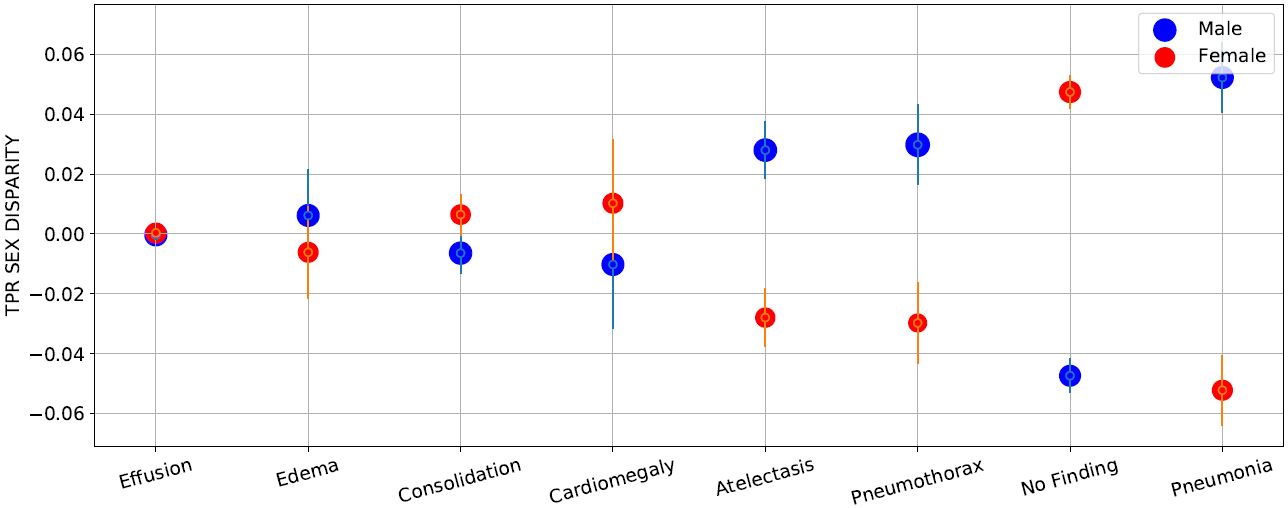}
      \caption{The sorted distribution of the TPR sex disparity in ALL dataset per disease. The $x$-axis labels are the disease names. The scatter plot's circle area is proportional to the patients percentages per subgroup. The TPR disparities are averaged over five run $\pm$95\% CI. The 95\% CI are shown with arrows around the TPR disparities mean scatter plot. The average cross labels gaps between the the least/most favorable subgroups is 0.045. Female are the most unfavorable subgroups with 4/7 count of negative disparities in disease labels where `Male' are the most favorable subgroups. Here, `Effusion' is the label with the smallest gap (0.001) between the least/most favorable subgroups, where `Pneumonia' has the largest gap (0.105).  }
       
      \label{Fig:GAPSeX1}

  \end{figure*}

\begin{figure*}[!htb]
    \centering
      \includegraphics[width=0.99\textwidth]{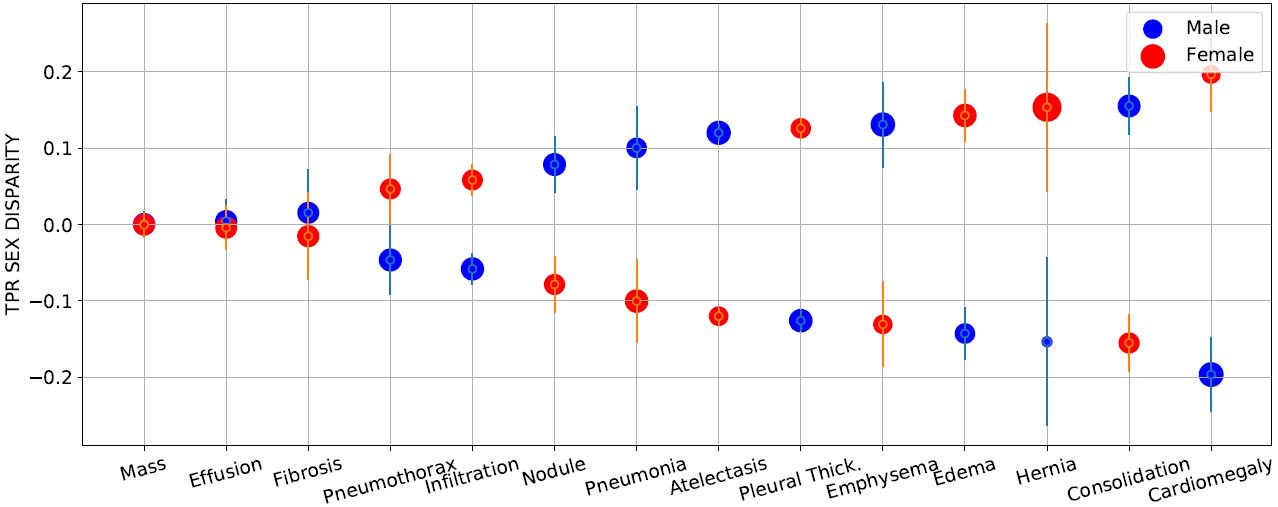}
      \caption{The sorted distribution of the TPR sex disparity in ChestXray8 dataset per disease. The $x$-axis labels are the disease names. The scatter plot's circle area is proportional to the patients percentages per subgroup.  The TPR disparities are averaged over five run $\pm$95\% CI. The 95\% CI are shown with arrows around the TPR disparities mean scatter plot. Count of `Female' and `Male' patients with negative disparities in disease labels are 8/14 and 6/14. Here, `Mass' (`M') is the label with the smallest gap (0.001) between the least/most favorable subgroups, where `Cardiomegaly' (`Cd') has the largest gap (0.393). The average cross labels gap between the the least/most favorable subgroups are 0.190. }

    \label{Fig:GAPSeX3}
  \end{figure*}

\begin{figure*}[!htb] 
    \centering
      \includegraphics[width=0.99\textwidth]{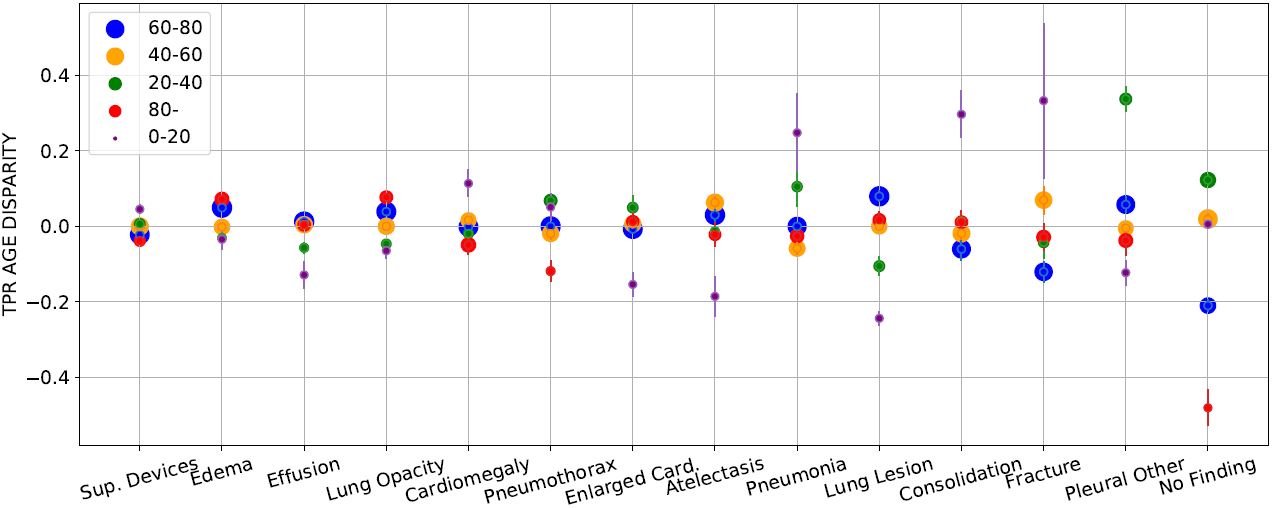}
      \caption{The sorted distribution of the TPR age disparity in CheXpert dataset per disease. The $x$-axis labels are the  disease labels. Here, the label Lung Opacity and Airspace Opacity label in Table 2 refer to the same disease. The scatter plot's circle area is proportional to the percentage of patients in each subgroup. The TPR disparities are averaged over five run $\pm$95\% CI (CI are shown with arrows around the mean). Count of patients in age subgroups `40-60', `60-80', `20-40',`80-' and `0-20'  with negative gap in disease labels are 5/13, 6/13, 7/13, 7/13 and 7/13. Here, `Support Devices' (`SD') is the label with the smallest gap (0.082) between the least/most favorable subgroups, where `No Finding' (`NF') has the largest gap (0.604). The average cross labels gap between the the least/most favorable subgroups are 0.270.}
      \label{AgeGapAllCXP}
  \end{figure*}

\begin{figure*}[!htb]
    \centering
      \includegraphics[width=0.99\textwidth]{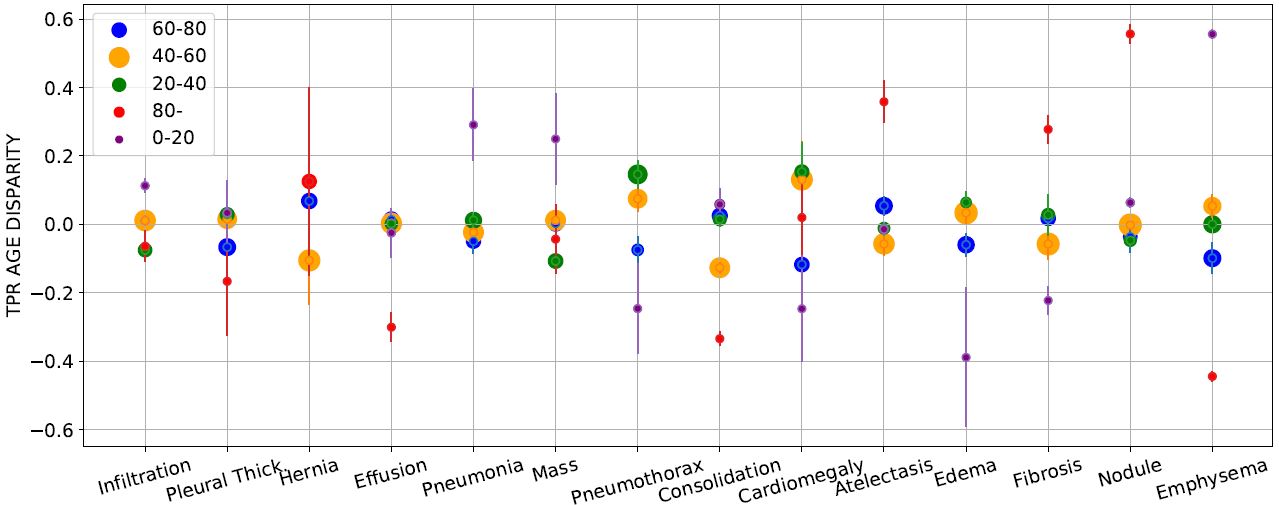}
      \caption{The sorted distribution of the TPR age disparity in ChestXray8 dataset per disease. The $x$-axis labels are the disease labels. The scatter plot's circle area is proportional to the patients membership. The TPR disparities are averaged over five run $\pm$95\%CI (the CI are shown with arrows around the mean). Count of patients in age subgroups `40-60', `60-80', `20-40',`80-' and `0-20'  with negative gap in disease labels are 6/14, 7/14, 4/14, 6/14 and 6/14. Here, `Infiltration' (`In') is the label with the smallest gap (0.188) between the least/most favorable subgroups, where `Emphysema' (`Em') has the largest gap (1.00). The average cross labels gap between the the least/most favorable subgroups are 0.413.}
      \label{AgeGapAllNIH}

  \end{figure*}

\begin{figure*}[!htb]
    \centering
      \includegraphics[width=0.99\textwidth]{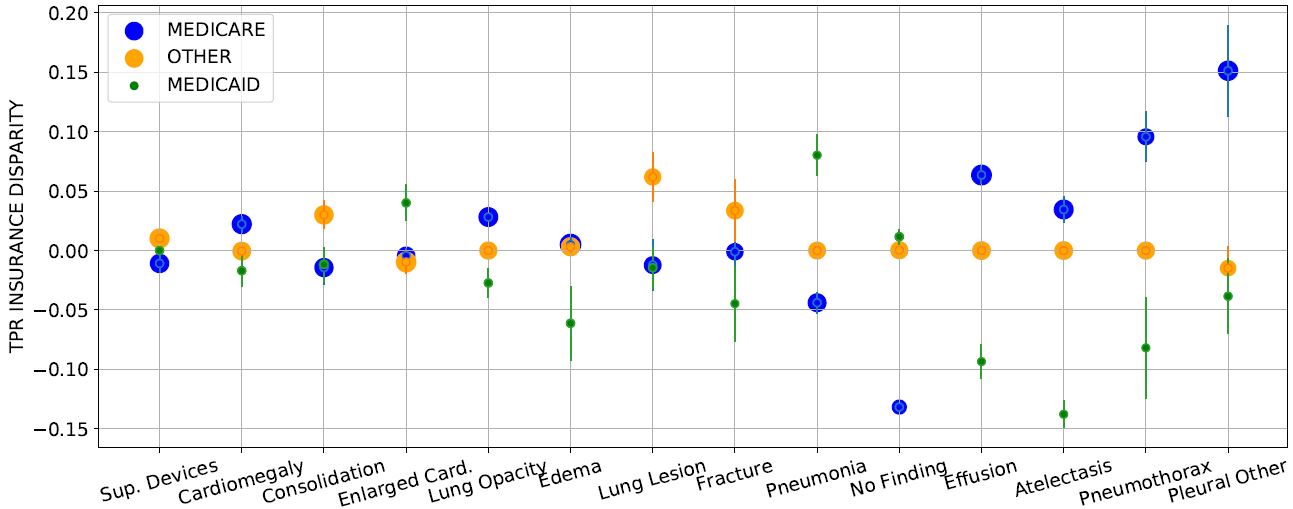}
      \caption{The sorted distribution of the TPR insurance type disparity in MIMIC-CXR dataset per disease.The $x$-axis labels are the disease names. Here, the label Lung Opacity and Airspace Opacity label in Table 2 refer to the same disease. The scatter plot's circle area is proportional to the patients membership. The TPR disparities are averaged over five run $\pm$95\%CI (the CI are shown with arrows around the mean).  Count of patients in insurance subgroups  `Other', `Medicare', and `Medicaid' with negative gap in disease labels are 3/13, 6/13, and 10/13. The patients with `Medicaid' insurance are the most unfavorable subgroup where `Other' are the most favorable subgroup with 10/13 positive disparity count. Here, `Support Devices' is the label with the smallest gap (0.021) between the least/most favorable subgroups, where `Pleural Other' has the largest gap (0.190). The average cross labels gap between the the least/most favorable subgroups are 0.100.}
      \label{MIMICInsAll}

  \end{figure*}

Here we present the distribution of TPR disparities per subgroups/disease labels for all attributes. In a fair setting all subgroups TPRs per disease are the same and disparity is `0'. Conversely, negative and positive disparities denotes bias against and in favor of a subgroup, respectively. The subgroup with largest (positive) and smallest (negative) TPR disparities per disease label are the most favorable and unfavorable subgroups, respectively.  In Fig. \ref{Fig:GAPSeX1} to Fig. \ref{MIMICInsAll}, we sort disease labels based on the gap between the least and most favorable subgroups per disease, so that ones with smaller variance in disparity appear on the left side.  We quantify TPR disparity across different subgroups similar to prior work \cite{de2019bias} for sex attributes, as the TPR of the subgroup of interest minus the TPR of the other subgroup (e.g.  $\operatorname{Disparity}_{Female, Edema}=\operatorname{TPR}_{Female, Edema}-\operatorname{TPR}_{Male, Edema}$). For age, race, and insurance type we quantify disparities using the difference between a subgroup's TPR and the TPRs median. We present the count of negative disparities per subgroup across all labels, excluding the `No Finding' (`NF') label in order to consider disease labels only.  The counts are based on the TPR disparities mean over five run. For Fig. \ref{Fig:GAPSeX1} to Fig. \ref{MIMICInsAll} the label with the smallest and largest gap (distance) between the least/most favorable subgroups, the average cross labels gaps (between the the least/most favorable subgroups), and the count of the most frequent `Unfavorable' and `Favorable' subgroups, are summarized in Table. 3 and presented in the figure captions.

\begin{figure*}[!htb]   
    \centering
      \includegraphics[width=0.99\textwidth]{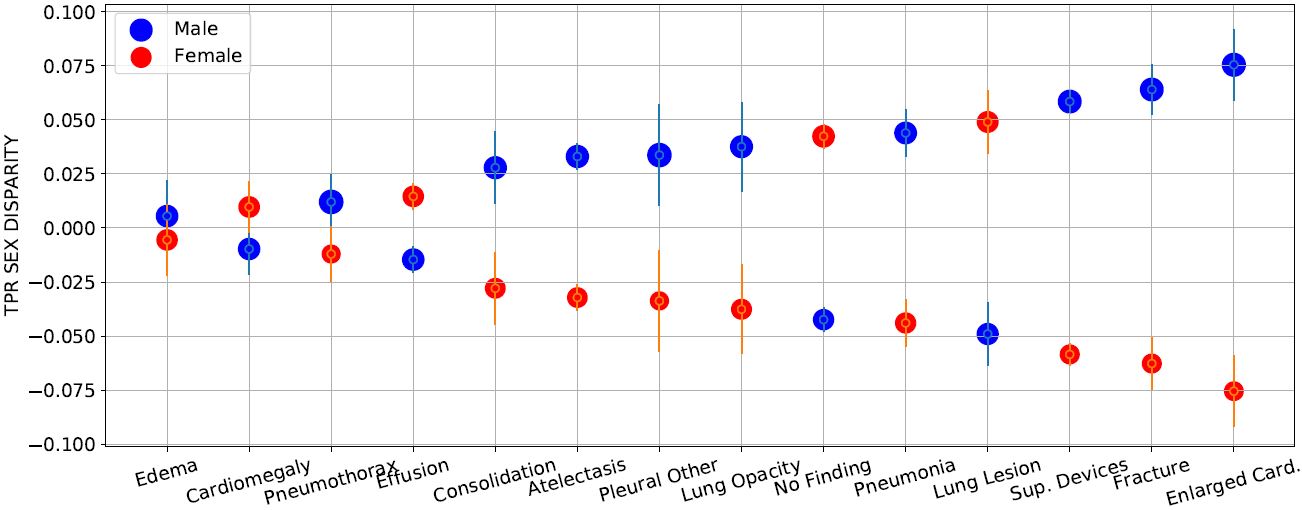}
      \caption{The sorted distribution of the TPR sex disparity in MIMIC-CXR dataset per disease. The $x$-axis labels are the abbreviation of the disease names. Here Lung Opacity and Airspace Opacity label in Table 2 refer to the same disease. The scatter plot's circle area is proportional to the patients percentages per subgroup. The TPR disparities are averaged over five run $\pm$95\% CI. The 95\% CI are shown with arrows around the TPR disparities mean scatter plot. Count of `Female' and `Male' patients with negative disparities in disease labels (excluding `No Finding') are 10/13 and 3/13. Thus Female are the most unfavorable subgroup. Here, `Edema' is the label with the smallest gap (0.011) between the least/most favorable subgroups, where `Enlarged Cardiomediastinum' has the largest gap (0.151). The average cross labels gap between the the least/most favorable subgroups are 0.072. }

  \end{figure*}

\begin{figure*}[!htb] 
    \centering
      \includegraphics[width=0.99\textwidth]{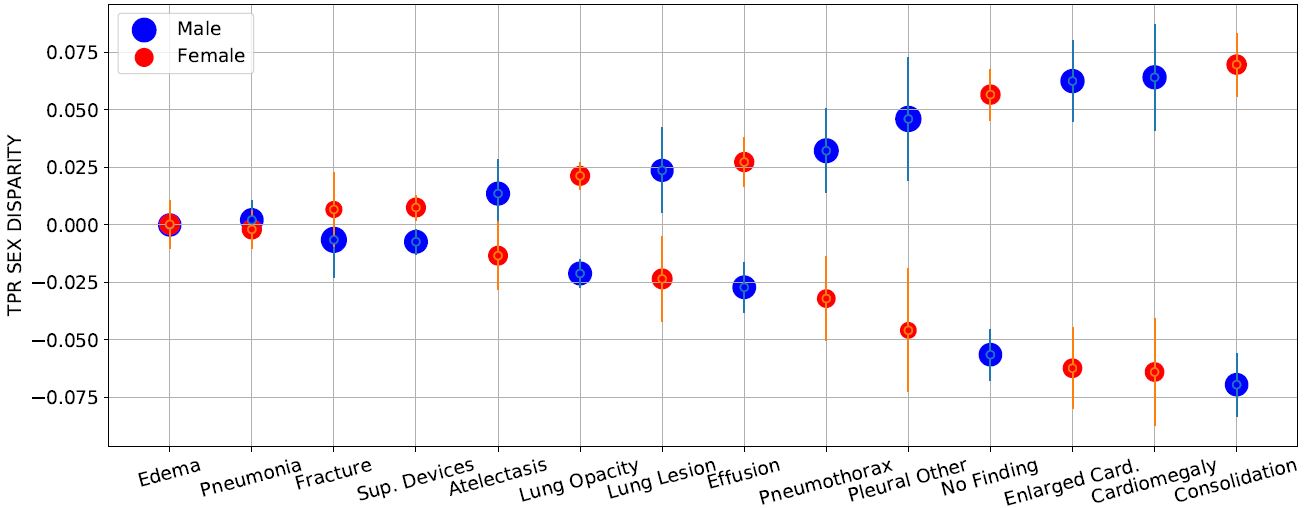}
       
      \caption{The sorted distribution of the TPR sex disparity in CheXpert dataset per disease. The $x$-axis labels are the disease labels. Here, the label Lung Opacity and Airspace Opacity label in Table 2 refer to the same disease. The scatter plot's circle area is proportional to the patients percentages per subgroup. The TPR disparities are averaged over five run $\pm$95\% CI. The 95\% CI are shown with arrows around the TPR disparities mean scatter plot. Count of `Female' and `Male' patients with negative disparities in disease labels are 7/13 and 6/13. Here, `Edema' (`Ed') is the label with the smallest gap (0.000) between the least/most favorable subgroups, where `Consolidation' (`Co') has the largest gap (0.139). The average cross labels gap between the the least/most favorable subgroups are 0.062. }
       
  \end{figure*}

\begin{figure*}[!htb]   
    \centering
      \includegraphics[width=0.99\textwidth]{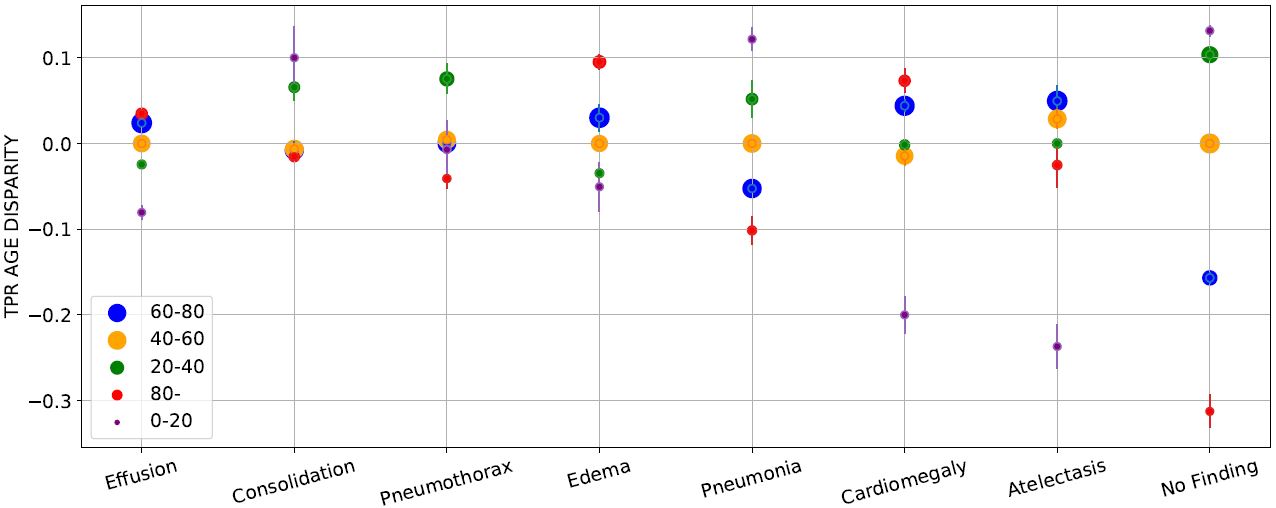}
      \caption{The sorted distribution of the TPR age disparity in ALL dataset per disease. The $x$-axis labels are the disease names. The scatter plot's circle area is proportional to the percentage of patients in each subgroup. The TPR disparities are averaged over five run $\pm$95\% CI. The 95\% CI are shown with arrows around the mean of TPRs scatter plot. The  count of patients in age subgroups `40-60', `60-80', `20-40',`80-' and `0-20'  with negative gap in disease labels are 2/7, 2/7, 4/7, 4/7 and 5/7. Thus ypung patients 0-20 are the most unfavorable subgroups where patients 40-60 and 60-80 are the most favorable subgroups with 5/7 count of positive gaps over disease labels. The average cross labels gaps between the the least/most favorable subgroups is 0.215. Here, `Effusion' is the label with the smallest gap (0.115) between the least/most favorable subgroups, where `No Finding' has the largest gap (0.444).}
      \label{AgeGapAllMIMIC}

  \end{figure*}

\begin{figure*}[!htb]   
    \centering
      \includegraphics[width=0.99\textwidth]{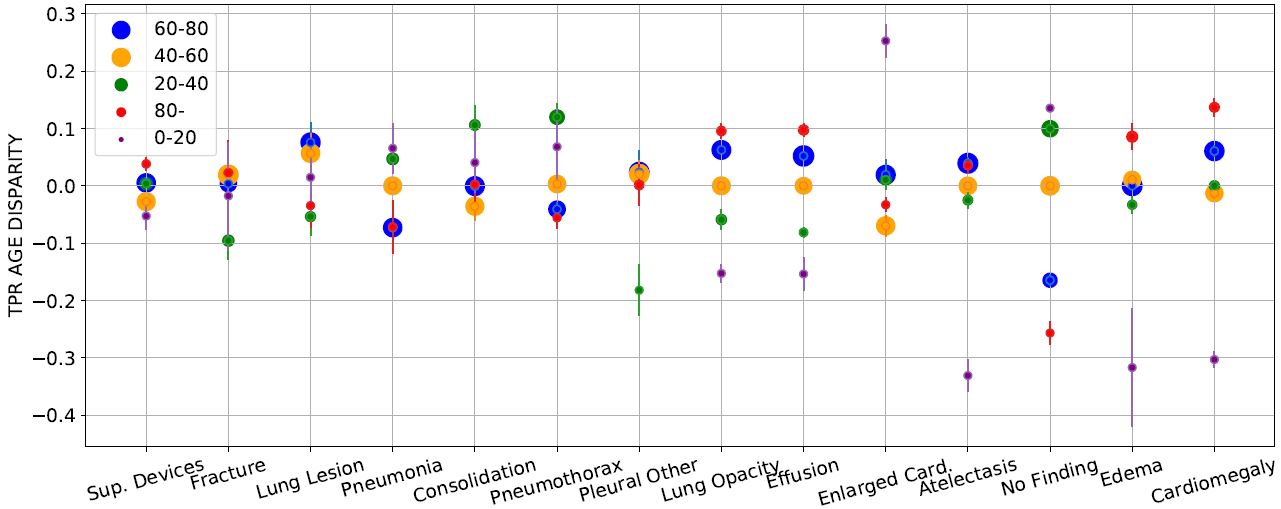}
      \caption{The sorted distribution of the TPR age disparity in MIMIC-CXR dataset per disease. The $x$-axis labels are the disease labels. Here, the label Lung Opacity and Airspace Opacity label in Table 2 refer to the same disease. The scatter plot's circle area is proportional to the percentage of patients in each subgroup. The TPR disparities are averaged over five run $\pm$95\% CI. The 95\% CI are shown with arrows around the mean of TPRs scatter plot. Count of patients in age subgroups `40-60', `60-80', `20-40',`80-' and `0-20'  with negative gap in disease labels are 4/13, 3/13, 7/13, 4/13 and 7/13. Thus, patients 0-20 and 20-40 are the most favorable subgroups where patient 60-80 with 10/13 positive disparities are the most favorable subgroup. Here, `Support Devices' is the label with the smallest gap (0.091) between the least/most favorable subgroups, where `Cardiomegaly' has the largest gap (0.440). The average cross labels gap between the the least/most favorable subgroups are 0.245.}
      \label{AgeGapAllMIMIC}

  \end{figure*}

\end{document}